\documentclass{article}

\usepackage[english]{babel}

\usepackage[a4paper,top=2cm,bottom=2cm,left=3cm,right=3cm,marginparwidth=1.75cm]{geometry}

\usepackage{amsmath}
\usepackage{authblk}
\usepackage{csquotes}
\usepackage{graphicx}
\usepackage{array}
\usepackage{float}
\usepackage[round]{natbib}
\usepackage[colorlinks=true, allcolors=blue]{hyperref}

\title{ABC Align: Large Language Model Alignment\\ for Safety \& Accuracy }
\author{Gareth Seneque}
\author{Lap-Hang Ho}
\author{Ariel Kuperman}
\author{\authorcr Nafise Erfanian Saeedi}
\author{Jeffrey Molendijk}
\affil{Australian Broadcasting Corporation}

\usepackage[skip=11pt plus1pt]{parskip}

\usepackage{enumitem}
\setlist[enumerate]{itemsep=3pt, topsep=0pt, parsep=0pt, partopsep=0pt}

\newcolumntype{+}{>{\global\let\currentrowstyle\relax}}
\newcolumntype{^}{>{\currentrowstyle}}
\newcommand{\rowstyle}[1]{\gdef\currentrowstyle{#1}%
#1\ignorespaces
}

\begin{document}

\maketitle

\begin{abstract}
\noindent
Alignment of Large Language Models (LLMs) remains an unsolved problem. Human preferences are highly distributed and can be captured at multiple levels of abstraction, from the individual to diverse populations. Organisational preferences, represented by standards and principles, are defined to mitigate reputational risk or meet legislative obligations. In this paper, we present ABC Align, a novel alignment methodology for LLMs that enables integration of the standards and preferences of a large media organisation into the LLM itself. We combine a set of data and methods that build on recent breakthroughs in synthetic data generation, preference optimisation, and post-training model quantisation. Our unified approach mitigates bias and improves accuracy, while preserving reasoning capability, as measured against standard benchmarks.
\end{abstract}

\section{Introduction}

\subsection{Overview}

In this paper, we present a novel approach to the alignment of Large Language Models (LLMs) we call ‘ABC Align’. This alignment is conducted in two settings: the fine-tuning of open-source models, and In-Context Learning (ICL) of closed-source ‘frontier’ models \citep{lin2023unlockingspellbasellms}. While the techniques and literature of In-Context Alignment (ICA) are less mature than fine-tuning based alignment, we achieve good results across both settings. By design, our methodology is applicable across any range of open- and closed-source models, offering provider independence and flexibility as underlying model capability grows with time. 

We use a variety of data that is both domain-specific and general. These data include news article content, organisational ‘AI Principles’ \citep{abc_ai_principles}, and human-reviewed question/answer pairs sourced from an internal Retrieval-Augmented Generation (RAG) tool. They are used as input to both generate datasets for fine-tuning, and in the case of the ‘AI Principles’, directly in the frontier model context window itself.

In the fine-tuning alignment setting, we use several techniques from the literature on synthetic data generation, knowledge distillation, and preference optimisation to produce several datasets conditioned on these data. We include results controlling for the impact of each of these methods and data across two open-source models, from Mistral AI and Meta respectively.

We see general performance improvements across industry standard benchmarks. The version of our dataset that combines all data and methods achieves a 23.51\% relative performance improvement on Meta’s Llama3-8B model on the TruthfulQA benchmark. Furthermore, these results were achieved using a small dataset of 897 samples, showing substantial improvement over Meta’s own fine-tuned version of the base model, which is trained on more than 10M human-annotated examples \citep{llama3modelcard}. 

For the ICA setting, conducted on OpenAI’s GPT4-turbo model, we evaluate the performance of a number of custom system prompts, including the prompt used by our internal RAG tool, and our AI Principles. Most notably in evaluation against the Bias Benchmark for Question Answering (BBQ), a system prompt that uses our AI Principles improves relative performance over baselines by 77.54\%.

Aligning LLMs and conducting evaluations are inherently challenging due to difficulty in defining ‘who the alignment is for’, and the need to define benchmarks reflecting a measure of alignment specific to that group, in addition to standard benchmarks. 

Our work aims to show an example of how organisations can answer these questions, integrate their standards into a process of provider-agnostic alignment, hence providing the organisations with choice and agency in how they apply LLMs across a range of use-cases.

\subsection{Background }

\textit{Authors’ note: no editorial position is offered or claimed in this publication.}\textit{\\We also do not publish our data, and all LLM prompts are provided to support replication of our work and do not reflect optimisation for production use.}

For real-world industrial application, in our case a large public-service media (PSM) organisation, alignment of model outputs to policies and preferences is critical. This is due to charter obligations to provide the public with innovative services \citep{abc_legislative} that mitigate reputational risk while being transparent. 

Many PSMs face additional constraints specific to their operating model, most notably in the link between their sources of funding and restrictions around commercial activity \citep{ebu_viewpoint}. This reality can limit their capacity to support internal R\&D efforts around technology, while they simultaneously rely on technology to deliver content to audiences in a highly competitive media marketplace.

LLMs present a unique opportunity for PSMs to improve internal processes. For example, through enhancing content metadata or supporting information retrieval across large corpora. LLMs also carry a unique risk profile \citep{llm_hallucination, llm_bias, llm_risks} when working with that content itself, whether it be to transform the content for multi-platform use, or use-cases that sit at the edge of emerging definitions around ‘responsible AI’, such as content generation \citep{ai_guidelines_global}.

Recent progress in model capability that enables processing of content across modalities in a single system – text, audio, video – only serves to enhance this risk profile and magnify impact of the constraints PSMs face, given the computational and operational costs associated with utilising such systems at scale.

Scale is a critical consideration for PSMs both in terms of the audience served and content produced and stored. Leveraging the capabilities of LLMs (and their multimodal derivatives) thus involves a considered trade-off between scale, cost, and safety. 

The Australian Broadcasting Corporation (ABC) has defined a set of AI Principles (see Appendix A) that are designed to guide the adoption and application of AI in a responsible, principled manner. We build on the use of ‘safety’ in the literature, particularly work on ‘Constitutional AI’ \citep{constitutional_ai}. We make explicit use of the ABC AI Principles in our methodology, and here include the principles of ‘protecting data’ and ‘mitigating bias’ to broaden the definition of ‘safety’. ‘Accuracy’ maps directly to the AI Principle of the same name. This paper outlines methods that securely leverage the ABC’s data and are evaluated against industry-standard benchmarks that act as proxies for these principles.

\subsection{ABC Align}

Our work sits at the intersection of machine learning theory and application. The organisational context in which it is conducted, under the various constraints outlined above, guides our choices of models, data and methods. These methods leverage the unique advantage of media organisations, namely their content, which is created in line with clearly defined policies and principles. We consider these as data valuable for the alignment and evaluation of LLMs, when paired with the approach we outline in this paper.

To best utilise this data, we build on recent work on synthetic data generation aimed at distilling frontier models’ capability into smaller open-source models using ICL \citep{distillation_survey}. Both the open-source and frontier models can be substituted, and the impacts of these changes are measurable. In this paper, we include representative results across several open-source models and a specific frontier model to validate and demonstrate our methods. Our aim is to offer a blueprint that is prescriptive on method only, enabling the replication and use of our work by other PSMs with their own unique data.

Our synthetic datasets, also referred to as augmented datasets, are used by a fine-tuning process that conditions each sample on textual article content, organisational ‘AI principles’, and samples gathered from the live deployment of an internally built assistive tool. This tool enables the capturing of editorial preferences and input at scale, hence also exploiting the ICL abilities of frontier models to guide and align the application to user needs. 

Development of this tool involved the crafting of novel, structured prompts and retrieval architectures, evaluated against user-derived acceptance criteria that were reflective of broader organizational preferences. We defer a lengthier discussion of its implementation to a future publication. 

In addition to our alignment methods in a fine-tuning setting, the results presented in this paper reflect nascent developments in ICA of frontier models, and we expect to build on these in future work, with more detail covered in our 'Future Work' section. We aim to provide a general methodology: general to our organisation’s data, the full complement of available AI models, and application to a range of use-cases specific to media organisations. Such applications can include metadata tagging, sourced information retrieval, and LLM-based approaches to content recommendation \citep{llm_rec}.

We leave the infrastructure enabling the synthetic dataset creation, knowledge distillation, training and evaluation of LLMs out of scope for this paper. We note that we make extensive use of open-source tools, including Hugging Face's Transformers and TRL \citep{wolf-etal-2020-transformers, vonwerra2022trl} libraries among others.

Evaluation also forms a key part of our methodology. We build on the work of EleutherAI, specifically their ‘Large Language Model Evaluation Harness’ \citep{lm_eval} and emerging best practices to measure our alignment efforts against industry-standard benchmarks that reflect the ‘AI Principles’ outlined above. The framework is extensible, enabling custom, organisation-specific evaluations defined in collaboration with non-technical stakeholders. We offer further commentary on this in the ‘Future Work’ section at the end of this paper.

\section{Related Work}

In this section we outline a brief history of LLM development, and each of the methods we use as part of ABC Align.

Early implementations of the ‘Transformer’ and related attention mechanism were limited in both scale of pre-training data and the number of parameters in the neural network \citep{attention}. However, the new architecture, with query-, key-, and value-cache to optimise both training and inference, went beyond the capacity of earlier architectures like Long Short-Term Memory (LSTM) and Recurrent Neural Networks (RNNs), both being limited in their ability to process long sequences under fixed computational constraints.

Radfort et al. \citep{gpt2_paper} moved beyond masked-language modelling approaches and described the Generative pre-trained Transformer (GPT) architecture, which framed the prediction problem in terms of arbitrary sequence-to-sequence modelling \citep{seq2seq, llm_few_shot_learners}. 

Ouyang et al. \citep{rlhf} directly improved this ability to follow general instructions via their method of ‘instruction tuning’, framing the model interaction as a ‘request/completion’ dynamic, leading to the conversational-style interfaces seen in recent iterations of GPT-class models.

As this approach was scaled up, the diversity and distribution of training data (consisting of trillions of tokens sourced from web-scale datasets), revealed limitations for the application of these models in a practical setting, where user expectations around accuracy and bias were not met \citep{askell_alignment}.

\subsection{Domain-Specific Large Language Models}

LLMs have been developed for application to specific fields or areas of industry, such as biology and finance \citep{ling2024domainspecializationkeymake}. A canonical example is BloombergGPT \citep{bloomberg_gpt}. Bloomberg has a significant store of domain-specific data (363B tokens) enabling it to perform full pre-training of their own model. Bloomberg’s experience informed the methodology outlined in this paper, as the ABC itself has a large store of data across digital-first content and its archives. 

A complete pre-training process of an ABC LLM is likely to yield a capable, general model, given that digital content alone contains over a billion tokens, and 90+ years of archival content is currently being digitised. However, the expense and risk associated with maintaining a single artifact given the rate of progress in the field \citep{ausgov_dprf_working_paper}, have led us to favour a lightweight, modular methodology rather than pre-training a base model from scratch. This would allow us to leverage the value of the ABC’s extensive data but in a way that was agnostic to underlying base models, in a cost-effective and future-proof manner. This has in part been validated by subsequent research, which has demonstrated that OpenAI’s GPT-4 has exceeded the performance of BloombergGPT \citep{li2023chatgptgpt4generalpurposesolvers}.

\subsection{LLM Alignment}

LLM alignment is a rapidly evolving area of research, alongside efforts to further scale up pre-training data and neural network parameter counts. Starting in 2023, early evidence \citep{lima} emerged that while massive amounts of data were required in the pre-training stage, alignment itself could be facilitated through smaller, high-quality curated datasets with samples numbering in the thousands instead of trillions. In parallel, LLM capabilities were being categorised relative to new evaluation frameworks, specifically around ‘reasoning’ (problem-solving) \citep{orca2}, extending standard definitions of accuracy beyond metrics like ROUGE and BLEU \citep{bleu}.

Our work builds on ‘Constitutional AI’ from Anthropic \citep{constitutional2} and ‘Less is More for Alignment’ (LIMA) by Meta \citep{lima}. Constitutional AI focuses on the self-alignment of LLMs based on a ‘constitution’ or set of principles that guide the process of ‘AI feedback’. We adapt this approach to use the ABC AI Principles directly in the creation of our fine-tuning datasets and frontier model alignment prompts. 

LIMA defines the ‘Superficial Alignment Hypothesis’, that states a \textit{‘model’s knowledge and capabilities are learnt almost entirely during pre-training, while alignment teaches it which sub-distribution of formats should be used when interacting with users’}. The use of ‘superficial’ here indicates that fine-tuning a subset of parameters, rather than complete pre-training, is the appropriate setting for LLM alignment. Their work examines the impact of a small set of 1000 samples on several alignment metrics, building on \citet{kirstain}. Our dataset is thus correspondingly small, high-quality, and informed by our AI Principles.

Our own hypothesis is that given a sub-distribution of fine-tuning and preference data specific to our organisation, we can in turn align LLMs in a way that preserves their general utility, capability, and performance on industry-standard benchmarks that directly map to and reflect our AI Principles. While non-domain-specific data like those from ‘Stack Exchange’, ‘Wikihow’ and ‘Reddit’ can be valuable in an academic fine-tuning setting, their use for alignment in a PSM setting is harder to justify. Consider a user feedback scenario, in the context of PSM obligations outlined in the introduction:

\begin{displayquote}
\textit{‘Why did your model generate this inaccurate/biased output?’\\
‘It was aligned using reddit threads!’}
\end{displayquote}
A key challenge that the LIMA authors observe is that \textit{‘manually creating diverse prompts and authoring rich responses in a uniform style is laborious’}. In our own organisational context, asking editorial staff to label data, even if for a small sample, represents a significant and resource-intensive task.

\subsection{Synthetic Data and Knowledge Distillation: ORCA}

The utilisation of ‘synthetic data’ \citep{synthetic_data_DL} represents a novel source of data for training LLMs. It is often cited in wider discussions about the finite quantity of human-created data \citep{villalobos2024rundatalimitsllm}. As with human data, not all synthetic data are created equal. Indeed, low-quality synthetic data, when riddled with hallucinations, inaccuracies, and biases, can significantly impact downstream task performance \citep{errors_synth_data}.

To improve the quality of our synthetic data and mitigate these issues, we condition each sample on organisational data: news articles. Beyond the content itself, these articles comply with the ABC’s editorial standards and policies \citep{abc_edpols} at the time of their publication, and provide a real-world example of ‘human annotated data’ that goes well beyond crowd-sourced methods such as or ‘Mechanical Turk’ or similar. We also go beyond algorithms like Self-Instruct \citep{self_instruct} that use ‘seed instructions’ to generate synthetic data, by providing a unique input for each generated sample.

Knowledge distillation techniques have diversified well beyond the original work of \citet{distill}. In the context of this paper, the teacher/student dynamic remains, but occurs between frontier model/smaller open-source model. The distillation occurs per-input, given a prompt. The structure of this prompt confines the possible sub-distribution of the synthetic data used by the student model in our fine-tuning process. Here we build on the work of \citet{orca} on ORCA, where they elicit ‘reasoning traces’ from OpenAI’s GPT-4 based on synthetic samples, with the goal of improving specific capabilities of a smaller 13B parameter model.

We note here that our synthetic data samples are created for the specific purpose of improving underlying model capability rather than meeting editorial standards, following the methods outlined in ORCA. We do not intend on making this data publicly available. This is to avoid any confusion with the published content it is derived from.

\subsection{Preference Optimisation: ORPO}

Soliciting human feedback for alignment efforts proved crucial in the success of tools like ChatGPT \citep{rlhf}. In a reinforcement learning setting, numerical instability and sensitivity to training dynamics were seen as a necessary trade-off for improved alignment performance. Model-free techniques such as Direct- and Odds-Ratio Preference Optimisation (DPO/ORPO) provide stable alternatives to the Reinforcement Learning with Human Feedback (RLHF) approach \citep{orpo}.

Following a pre-training stage, models undergo further post-training to make them suitable for general use. Multiple techniques exist to facilitate this, with initial work focused on RLHF \citep{ziegler_ft, stiennon_summarize_feedback}. This method aims to use human preference data to align pre-trained models and has demonstrated improved results beyond safety and reducing harmfulness \citep{tian_ft_factuality, gorbatovski_rlqa}, which are the key areas of focus for our work.

Recent work on preference optimisation \citep{dpo} aims to elicit similar changes in model behaviour via methods that reduce implementation complexity compared to RLHF. Given the constraints on PSMs (and organisations whose operational focus is not technology itself), these methods provide a way to demonstrate the value of incorporating their own ‘human feedback’ in the alignment of models they use.

In this paper, we make use of an implementation of ORPO that is readily available in Hugging Face’s libraries. ORPO offers good performance when compared with both Supervised Fine-Tuning (SFT) and DPO, including improved computational efficiency for a given batch of training data, making it suitable for the aims of our work.

\subsection{In-Context Alignment}

There has been significant recent work on exploiting the ICL capabilities of frontier models \citep{manyshotICL} where the context length has increased by up to an order-of-magnitude, as seen for instance in Google’s Gemini and Anthropic’s Claude series. While performance of fine-tuning in an ICL setting doesn’t yet offer consistent improvements when compared with SFT-based approaches across all tasks, there are specific cases where results are promising \citep{icl_vs_sft}. 

The work of \cite{lin2023unlockingspellbasellms} further quantifies the ‘Superficial Alignment Hypothesis’ proposed in LIMA. They measure the token distribution shift between pre-trained and aligned models, specifically the difference in probabilities of predicted tokens given a position in some sample input across the two models. They find that alignment fine-tuning affects a small subset of tokens that they describe as ‘stylistic’, specifically ‘discourse markers, transitional words, and safety disclaimers’. 

Further to the original hypothesis of the LIMA authors, this provides evidence and motivation for the effectiveness of alignment happening in a post-training setting via fine-tuning, and that inference-time alignment may offer comparable results for applications where fine-tuning is unsuitable or not possible. This further validates our decision to not pre-train an LLM, but instead focus on post-training alignment across SFT, PO, and ICA settings in support of our overall aim to provide a lightweight, flexible methodology that considers organisational constraints and the rapid growth of underlying model capabilities.

\subsection{Model Quantisation}

As we discussed in the introduction of this paper, PSMs are funding-constrained, thus any training of LLMs needs to be done in an efficient way. Early methods of model quantisation involved performance trade-offs in terms of prediction quality and performance when compared to full-precision training and inference. However, recent advances in model quantisation have enabled rapid and low-cost experimentation. QLoRA \citep{qlora} addresses the shortcomings of earlier methods sufficiently and is our choice for quantisation for the experiments outlined in this paper.

\subsection{Retrieval Augmented Generation \& Grounding Synthetic Data}

Another key aspect of the ABC Align methodology is the refining of the preference dataset, using user-generated outputs from an internally developed RAG tool. This tool is powered by OpenAI’s GPT-4-32k \citep{gpt4_techreport} and uses context information retrieved from a corpus of ABC documents to answer user questions. 

Within this context, alignment of the frontier GPT-4 model refers to ensuring that the entire RAG tool produced outputs that were relevant, accurate, and valuable to end users. This ultimately served to reflect whether the application (and thereby the frontier model) met users' expectations, and by proxy their broader organisational values. This effort involved a comprehensive system-wide approach, including not only refining the prompting techniques but also ensuring that the retrieval mechanisms, data sources and generated outputs all complied with the overall expected objectives \citep{rag_survey}. Evaluation of the system includes using user-acceptance derived metrics, conducting user interviews, and manually reviewing generated samples, ensuring an adequate level of alignment of the frontier GPT-4 model to the ABC's AI Principles.

While we generate ORCA-style synthetic data in this work, RAG grounding of synthetic data to produce reasoning traces that are accurate to, rather than derived from, content data will be explored in future work.

\subsection{Information Theoretic Measures in NLP}

A primary focus of this paper is dataset quality, and we demonstrate its impact on LLM performance in multiple fine-tuning settings. While measurement against standard benchmarks is critical, we are also interested in measuring the quality of the data itself, as it is initially transformed during our synthetic data and knowledge distillation process. Information theoretic metrics offer a principled approach to measuring quality and changes in the information that the data represents. For this purpose, we use several well-known metrics, including Shannon entropy, mutual information, and Kullback-Leibler divergence \citep{info_theory}.

\section{Methodology}
\subsection{Synthetic Data Generation}

In this section, we outline two versions of our alignment dataset for full fine-tuning of open-source models, together with the synthetic dataset generation prompts. The inputs to these datasets are news articles, the ABC AI Principles, and samples from our internal RAG tool. 

This tool, while an early prototype, is designed to deliver the capabilities of LLMs directly to users while also providing a means for soliciting editorial feedback and preferences, key to scaling our methods, outlined in the ‘Future Work’ section of this paper.

An automated pipeline was developed to enable the creation of synthetic datasets by augmenting existing data, as outlined in the previous section. The step-by-step process of creating an ABC Align dataset suitable for SFT is as follows:

\begin{enumerate}
    \item Collect news articles as the input.
    \item Create LLM-generated question/answer pairs relating to each input news article using an ORCA-style prompt.
    \begin{enumerate}
        \item Prompt: \textit{Deduce any reasoning or logical problems from this article: }\texttt{\{context\}} \textit{Generate a complex question and a logical answer that requires step-by-step thinking, and elaborate on this thinking process as part of the answer. Both the question and answer must not refer to the original article. }
    \end{enumerate}
\end{enumerate}

The resulting dataset is then used to create a dataset suitable for preference optimisation (PO). The answer for each sample is rewritten considering the AI Principles, given two canonical examples of human-reviewed Q\&A pairs from our internal RAG tool.

\begin{enumerate}
    \item Rewrite the answers from the SFT dataset
        \item For ‘chosen’ answers:
        \begin{enumerate}
            \item Prompt: \textit{Here are two high-quality, human-reviewed Question/Answer pairs: Question:} \texttt{\{q1\}} \textit{Answer:} \texttt{\{a1\}} \textit{Question:} \texttt{\{q2\}} \textit{Answer:} \texttt{\{a2\}} \textit{Now review a new Question \& Answer pair: }\texttt{\{context\}} -- \textit{Rewrite the answer to better align with these principles:} \texttt{\{abc\_ai\_principles\}} \textit{Do not reference the principles in your response. Include only the text of the answer.’}
        \end{enumerate}
        \item For ‘rejected’ answers:
        \begin{enumerate}
            \item Prompt: \textit{Here are two high-quality, human-reviewed Question/Answer pairs: Question:} \texttt{\{q1\}} \textit{Answer:} \texttt{\{a1\}} \textit{Question:} \texttt{\{q2\}} \textit{Answer:} \texttt{\{a2\}} \textit{Now review a new Question \& Answer pair: }\texttt{\{context\}} -- \textit{Rewrite the answer to be unaligned with these principles:} \texttt{\{abc\_ai\_principles\}} \textit{Do not reference the principles in your response. Include only the text of the answer.’}
        \end{enumerate}
    \end{enumerate}

In all cases, we use OpenAI’s GPT-4-turbo frontier model with a temperature of 0.7 to generate synthetic data for use in a fine-tuning setting. The resulting datasets comprise 897 individual samples.

\subsection{Control Datasets}

We include three additional datasets in our experiments. These are OpenORCA \citep{OpenOrca}, LIMA in the SFT setting \citep{lima_dataset}, and UltraFeedback \citep{ultrafeedback} for ORPO training.

OpenORCA is a replication of the original ORCA data which was not made public. The dataset is sourced from FLAN-T5 and generated using both OpenAI’s GPT-3.5-turbo and GPT-4 models. The authors aim to reflect the distribution of the dataset outlined in the ORCA paper.

UltraFeedback is an updated version of the dataset used in the ORPO paper that addresses a data contamination issue related to the TruthfulQA benchmark.

The LIMA dataset is the original as published by the authors, and in all cases we randomly take 897 samples for direct comparison with our own datasets.

\subsection{Models and Fine-tuning}

We fine-tuned the following pre-trained models: Llama-3-8B \citep{llama3modelcard} and Mistral-7B-v0.3 \citep{mistral7b}, including both the base and instruction-tuned models. We used the default tokenizer of each of the models, and applied Hugging Face’s Zephyr-style chat template (with \texttt{<|system|>
}, \texttt{<|user|>}, and \texttt{<|assistant|>} tokens).

As described above, we employed a two-step fine-tuning process involving SFT for instruction tuning and ORPO for preference optimisation. To facilitate this, we used TRL’s SFTTrainer and ORPOTrainer \citep{vonwerra2022trl} and scripts from Hugging Face’s Alignment Handbook \citep{alignment_handbook2023}.

ORPO includes an additional training objective, relative ratio loss. It also dynamically penalises ‘disfavoured’ responses, instead of constructing sets of rejected tokens to control for degenerative model behaviour in the SFT setting. 

We trained on single Nvidia A10G GPUs with 24GB memory on cloud infrastructure. We trained for 23 epochs on the SFT dataset, taking around 2.4 hours, and we trained for 10 epochs on the ORPO dataset, taking around 3.8 hours, using a standard checkpoint selection process against selected benchmarks for the models used in our final evaluations.

\subsection{QLoRA}

The SFT/ORPO QLoRA hyperparameters used in this study followed those as set in the Hugging Face Alignment Handbook. An overview of our QLoRA hyperparameters during model fine-tuning are shown in Table \ref{tab:qlora}.

\begin{table}[H]
    \centering
    \begin{tabular}{|+c|^c|^c|^c|^c|} \hline  
    \rowstyle{\bfseries}
 Configuration& Dropout& Rank&Alpha &Rank/Alpha Ratio\\ \hline  
         SFT&  0.1&  64&  16&4\\ \hline  
         ORPO&  0.05&  16&  32&0.5\\\hline 
    \end{tabular}
\caption{QLoRA hyperparameters used for fine-tuning.}
\label{tab:qlora}
\end{table}
Hyperparameter selection for quantisation methods has been shown to influence model performance in numerous ways, where the model may have different optimal ‘rank’ hyperparameters for different metrics, as well as the ratio between alpha/rank \citep{lora}.

The rank hyperparameter is particularly important since it affects the size of the update matrices, and therefore the number of trainable parameters. It should be noted that these recommendations are evolving, and hyperparameter optimisation may be required for individual use-cases \citep{hyperparam_opt_instruction_tuning}.

\section{In-Context Alignment}

 For our experiments on ICA, we use OpenAI’s GPT-4-turbo model with a temperature of 0.7, and the \texttt{lm-evaluation-harness} evaluation framework. Benchmarking is done in a `\texttt{generate\_until}' setting rather than traditional multiple choice, due to the limitations around the availability of logits from closed-source APIs.

 Our primary intervention is setting the `\texttt{system\_instruction}' parameter, including three specific prompts, taken directly from our internal RAG tool, the ABC’s AI Principles, and a version of the AI Principles augmented with a Q\&A sample sourced from the RAG tool, acting as a canonical example.

\subsection{Dataset Analysis }

This section outlines the information-theoretic and perplexity metrics we used to evaluate the datasets discussed in this study.

\subsection{Shannon Entropy, Mutual Information, KL Divergence}

 For dataset analysis, we utilise several information-theoretic metrics on our SFT dataset. Our motivation here is to quantify the changes to average entropy, loss of information, and distribution shift induced by our synthetic data generation process. Our aim is to understand how our methods transform the information present in our input data, as we distil ‘reasoning traces’ suitable for downstream SFT and ORPO fine-tuning. All metrics are normalised.

In Table \ref{tab:shannon_entropy}, we consider the average Shannon entropy of our input news articles, and the resulting synthetic dataset.

\begin{table}[H]
    \centering
    \begin{tabular}{|+c|^c|^c|} \hline  
    \rowstyle{\bfseries}
        Dataset& Avg. Shannon Entropy& Std Dev.\\ \hline  
         News Articles (NA) &  0.754&  0.021\\ \hline  
         ABC Align SFT &  0.813 &  0.019\\\hline 
    \end{tabular}
\caption{Shannon Entropy across input news article dataset and the ABC Align dataset.}
\label{tab:shannon_entropy}
\end{table}

In Table \ref{tab:kl_div}, we calculate both mutual information and KL divergence across both datasets.

\begin{table}[H]
    \centering
    \begin{tabular}{|+c|^c|^c|} \hline  
    \rowstyle{\bfseries}
        Datasets& Mutual Information& KL-Divergence \\ \hline  
         NA/SFT &  0.142&  0.049\\ \hline  
    \end{tabular}
\caption{Mutual Information and KL-Divergence between the news article and the synthetic dataset generated using ABC Align.}
\label{tab:kl_div}
\end{table}

These results demonstrate that, through our process of conditioning synthetic data samples on news articles, we have increased sample complexity relative to the input while keeping shared information low (in-line with our attempt to derive specific patterns with high complexity and abstraction, i.e. ‘reasoning traces’). We have been able to do this while preserving the similarity of the overall distributions, as reflected in the low KL-Divergence score.

\subsection{Perplexity}
For perplexity (ppl), we measure the mean ppl per dataset per base model. We calculate the perplexity of all datasets for both pre-trained Llama-3-8B and Mistral-7B-v0.3 models. These metrics are shown in Table \ref{tab:ppl} for the ABC Align, LIMA, OpenORCA and UltraFeedback datasets. We note that the ORPO datasets include `chosen' and `rejected' formats, reflecting their preference optimisation nature in line with Hugging Face's Alignment Handbook \citep[Fine-tuning on your datasets $\vert$ DPO and ORPO]{alignment_handbook2023}. For the ABC Align ORPO dataset, these were generated by OpenAI’s GPT-4-turbo frontier model to be aligned/unaligned to the ABC AI Principles. 

\begin{table}[H]
\begin{center}
\begin{tabular}{|+c|^c|^c|^c|}
\hline
\rowstyle{\bfseries}
Dataset & Format & Llama-3-8B & Mistral-7B-v0.3 \\ \hline
ABC Align SFT & messages & 10.25 & 10.54 \\ \hline 
ABC Align ORPO & chosen & 9.37 & 9.10 \\ \hline 
ABC Align ORPO & rejected & 10.04 & 10.54 \\ \hline 
LIMA & messages & 10.04 & 11.75 \\ \hline 
OpenOrca & messages & 9.43 & 15.06 \\ \hline 
UltraFeedback & chosen & 6.23 & 6.83 \\ \hline 
UltraFeedback & rejected & 7.49 & 11.76 \\ \hline
\end{tabular}
\end{center}
\caption{Perplexity (ppl) of all training datasets for pre-trained LLama-3-8B and Mistral-7B-v0.3.}
\label{tab:ppl}
\end{table}

Interestingly, among the ORPO datasets, the ‘chosen’ format shows lower perplexity than the ‘rejected’ format. Rather than exhibiting equal perplexity for the chosen/rejected formats, this implies that the original model training may have achieved similar results as we aimed to achieve by aligning using the ABC AI Principles. 

We limit our use of ppl to the datasets themselves, rather than the fine-tuning phase. As observed by the LIMA authors, there is an anti-correlation between perplexity measured at training time and generational quality; perplexity can increase without generation quality necessarily decreasing \citep[Appendix B]{lima}. 

Finally, we note that while details of the pre-training datasets of either base model have not been released, data quality is known to impact pre-training performance \citep{pretrainers_guide_data}, and indeed Meta does emphasise the use of filtering techniques to select only ‘high-quality’ data \citep{llama3_blogpost}.

\section{Model Evaluation}

 In this section we discuss our evaluation strategy, including the mapping between selected benchmarks and the ABC AI Principles. We evaluate both our datasets and model outputs, whereby results are presented in the section below.

\subsection{General Reasoning}

In the fine-tuning setting, we want to ensure that we are not training the model to be task-specific. Here we use the ARC dataset, specifically the ‘challenge’ subset, constructed by the Allen AI Institute \citep{arc_challenge}. It comprises 7,787 science questions in a multiple-choice setting aimed towards measuring ‘reasoning capability’. We select this benchmark to ensure that we do not sacrifice general model capabilities while improving bias identification and accuracy, i.e., limiting their application across a range of use-cases – also known as the ‘alignment tax’ \citep{askell_alignment}.

\subsection{ABC AI Principle: ‘mitigating bias’}

The Bias Benchmark for Question \& Answering (BBQ) \citep{bbq} represents a human-constructed and thorough cross-section of socio-cultural biases in a multiple-choice setting, for the evaluation of a model’s ability to detect bias. In this study, we use \texttt{bbq\_lite\_json}, a subset version of the original BBQ, containing samples for evaluation against age, nationality, race, gender, sexual orientation, and physical appearance biases. 

This benchmark serves as a proxy for the ABC AI Principle ‘mitigating bias’, which is defined as \textit{‘We will seek to prevent bias in our AI data or algorithms that could perpetuate and amplify existing inequalities and lead to discrimination’}. BBQ is designed specifically to measure whether model outputs contribute to ‘harming marginalised individuals and groups’, thus enabling the identification and mitigation of bias. The list of social biases measured are defined by the US Equal Employment Opportunity Commission \citep{us_employment_commission}. We acknowledge the US-centricity of this definition, and outline how we intend to extend our evaluation metrics in a general way, to make them ABC- or other-specific in our ‘Future work’ section at the end of this paper.

\subsection{ABC AI Principle: ‘accuracy’}

We use performance on TruthfulQA \citep{truthfulqa} as a proxy for the ABC AI Principle ‘accuracy’. TruthfulQA consists of 817 multiple-choice questions across 38 subjects, including law, health, and politics. This benchmark addresses the limitations of LLMs, where the distribution of their pre-training data may induce the generation of ‘imitative falsehoods’, or answers that sound true without being grounded. TruthfulQA thus represents a robust benchmark for accuracy, where the conditions for a claim to be true are \textit{‘if it describes..[a]..literal truth about the real world’}, and the question/answer choices are written by humans. We specifically use the \texttt{truthfulqa\_mc2} task.

\section{Results}
In this section we present the results of our fine-tuning alignment using both SFT and PO datasets.

\subsection{Fine-tuning: SFT (ORCA/LIMA)}

In an SFT setting, we evaluate the performance of our ABC Align SFT dataset against data generated using the two methods we build on, ORCA and LIMA respectively. We present our evaluation results in Figure 
\ref{fig:sft_results} below.

\begin{figure}[H]
    \centering
    \makebox[\textwidth][c]{\includegraphics[width=0.8\textwidth]{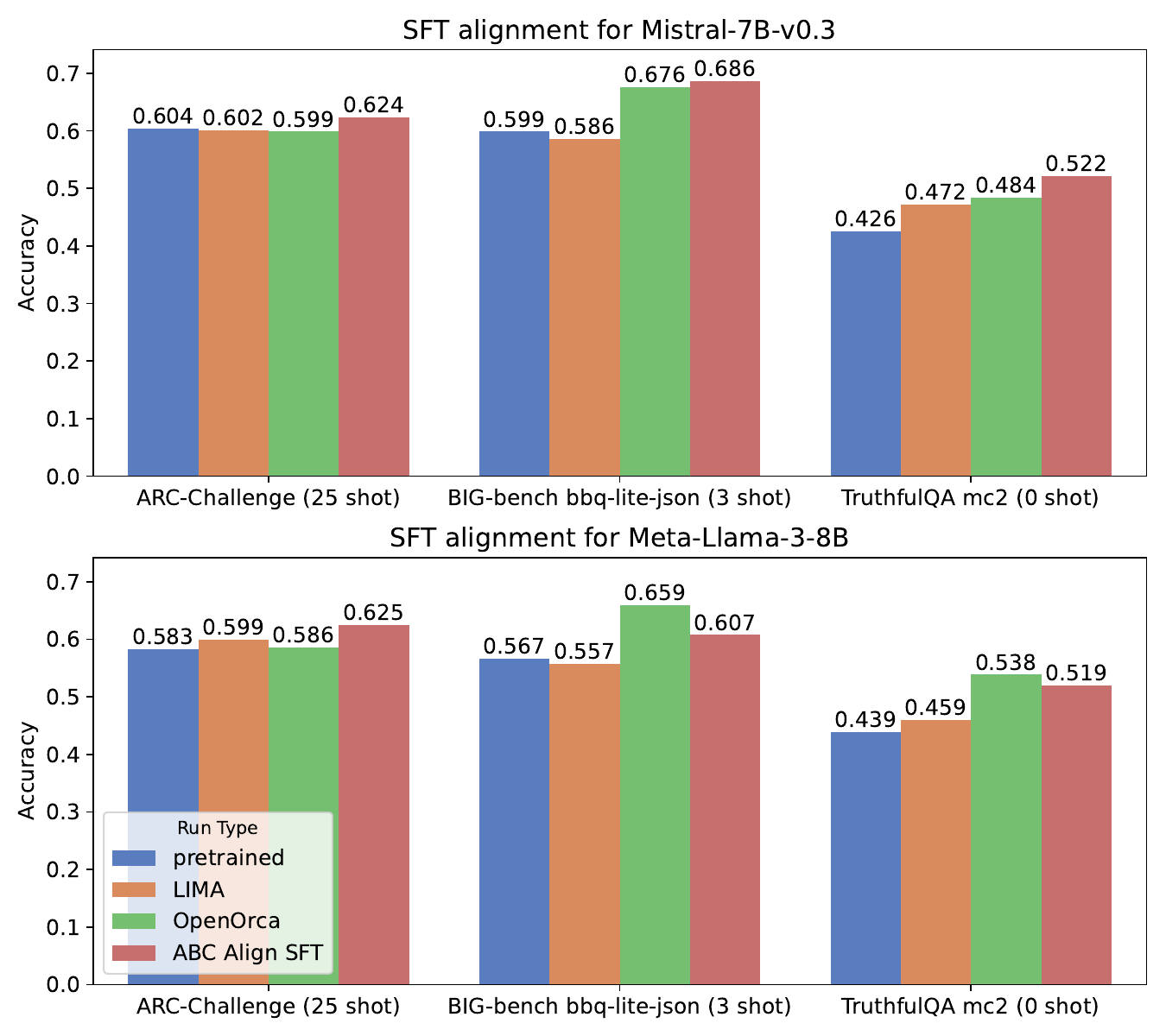}}
    \caption{Comparison of ABC Align SFT dataset against other methods, across \texttt{arc-challenge} (left), \texttt{bbq-lite-json} (middle) and \texttt{truthfulqa\_mc2} (right).}
    \label{fig:sft_results}
\end{figure}

\subsubsection{Evaluation on \texttt{arc-challenge}}

For Mistral-7B-v0.3, we see a decrease in performance compared to the pre-trained model of 0.42\% and 0.85\% for LIMA and OpenORCA. The Align SFT dataset demonstrates a 3.25\% improvement compared to the pre-trained model and those fine-tuned on control datasets. 

For Llama-3-8B, we see improvements over the pre-trained model for LIMA and OpenORCA datasets of 2.78\% and 0.59\% respectively. The ABC Align dataset demonstrates a 7.17\% improvement over the pre-trained baseline.

\subsubsection{Evaluation on \texttt{bbq-lite}}

For Mistral-7B-v0.3, we see a decrease in performance compared to the pre-trained model of 2.17\% on the LIMA dataset. For OpenORCA and ABC Align SFT, we see improvements of 12.81\% and 14.57\% respectively.

For Llama-3-8B, we see an equivalent decrease in performance of LIMA against the pre-trained model of 1.75\%. For OpenORCA and ABC Align SFT, we see improvements of 16.34\% and 7.20\% respectively over the pre-trained baseline.

\subsubsection{Evaluation on \texttt{truthfulqa\_mc2}}

For Mistral-7B-v0.3, OpenORCA and LIMA improve over pre-trained baselines by 11.01\% and 13.69\% respectively. ABC Align SFT demonstrates an improvement of 22.64\%.

For Llama-3-8B, LIMA improves over the baseline model by 4.56\%. OpenORCA and ABC Align improve by 22.57\% and 18.26\% respectively.

\subsubsection{Analysis}

In general, LIMA underperforms in this setting. ABC Align SFT offers the best performance on Mistral-7B-v0.3 across all benchmarks, and on Llama-3-8B for \texttt{arc-challenge}. OpenORCA’s performance exceeds ABC Align SFT on BBQ and TruthfulQA on Llama3-8B.

We see our dataset offers consistent improvements on a specific model, Mistral 7B. Where it underperforms on BBQ \& TruthfulQA on Llama-3, it has improved general reasoning capability. The SFT dataset is generated with this aim in mind, as reflected in the ORCA-style prompt. Ideally, we would like to see consistent improvements across a range of open-source models and our three evaluations. These results demonstrate that we need to move beyond SFT and explore PO to achieve this.

\subsection{Fine-tuning: PO (ORPO)}

The PO setting represents the ideal setting for the aims of this paper, where for the ABC Align dataset we have used each of the sources of data unique to our organisation. Preference data in particular is scalable beyond the small number of samples used via our internal RAG tool, to include data sourced from additional workflow integrations. 

\begin{figure}[H]
    \centering
    \makebox[\textwidth][c]{\includegraphics[width=0.8\textwidth]{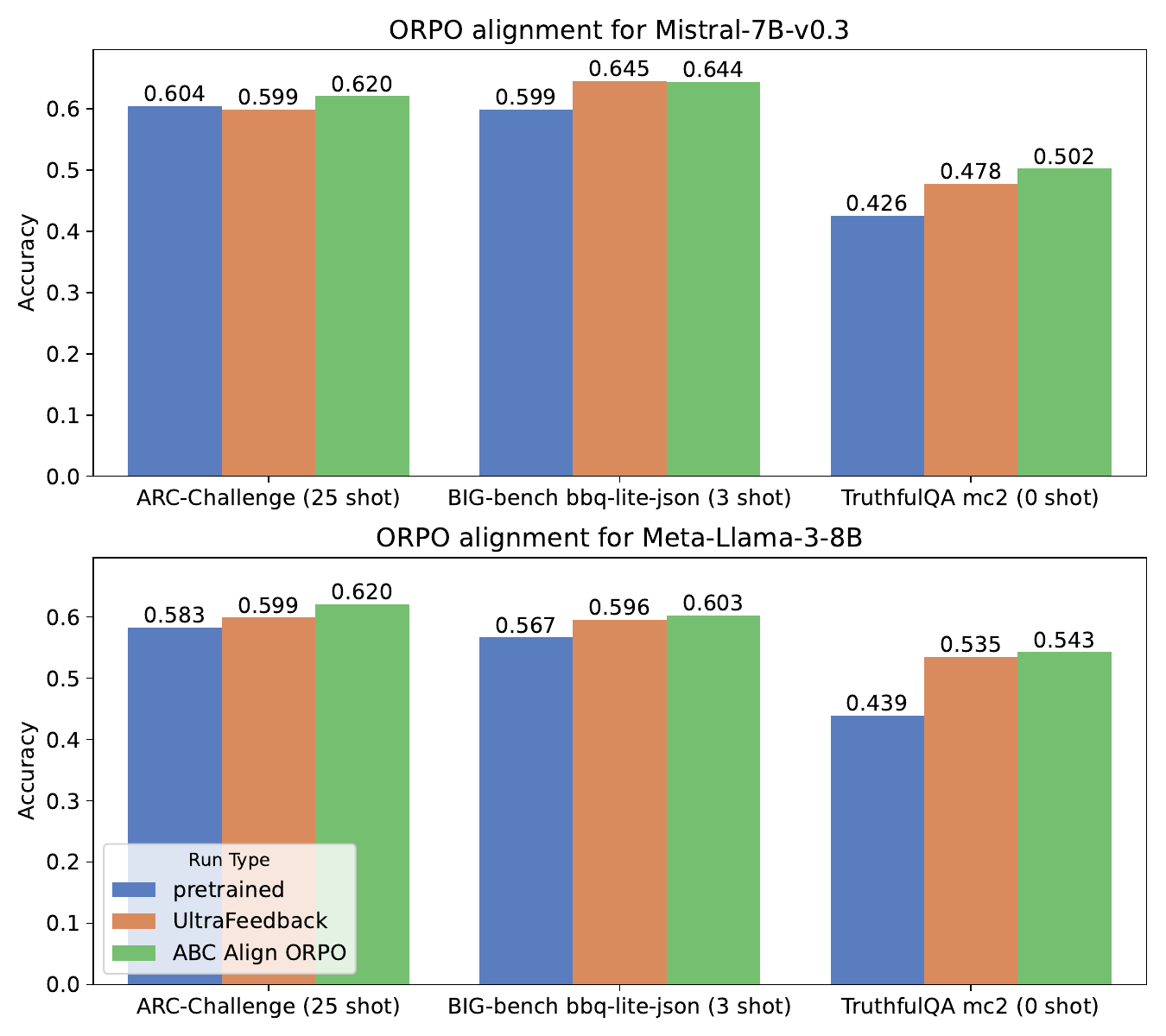}}
    \caption{Comparison of ABC Align ORPO dataset against control datasets, across \texttt{arc-challenge} (left), \texttt{bbq-lite-json} (middle) and \texttt{truthfulqa\_mc2} (right).}
    \label{fig:orpo_results}
\end{figure}

\subsubsection{Evaluation on \texttt{arc-challenge}}

For Mistral-7B-v0.3, performance on the pre-trained model decreases by 0.85\% for the UltraFeedback (UF) dataset, and ABC Align PO offers a 2.68\% improvement. 

For Llama-3-8B, UF and ABC Align PO offer improvements of 2.78\% and 6.44\% respectively compared to the pre-trained model.

\subsubsection{Evaluation on \texttt{bbq-lite}}

For Mistral-7B-v0.3, UF and ABC Align PO both improve over the pre-trained baseline by 5.26\%.

For Llama-3-8B, UF and ABC Align PO both improve over the pre-trained baseline by 7.64\% and 7.48\% respectively.

\subsubsection{Evaluation on \texttt{truthfulqa\_mc2}}

For Mistral-7B-v0.3,, we see improvements over the pre-trained baseline for both UF and ABC Align PO of 12.25\% and 17.97\% respectively.

For Llama-3-8B, we see improvements over the pre-trained baseline for both UF and ABC Align PO of 21.75\% and 23.51\% respectively.

\subsubsection{Analysis}

Here we see ABC Align PO outperforms or offers equivalent performance to the control dataset used by ORPO in all cases. These results demonstrate consistent performance improvements across both base models and all benchmarks. They emphasise the importance of including additional data sources in a PO setting, where ORPO’s penalisation of disfavoured responses better aligns model outputs in a measurable way. 

\subsection{Comparison of Instruction Fine-Tuned (IFT) and SFT/PO Align models}

In this section we include results comparing our models and the Instruction Fine-Tuned (IFT) variants of both Mistral-7B-v0.3 and Llama3-8B. Our aim in performing this evaluation is to quantify the impact of large datasets in a fine-tuning setting on the models and benchmarks we have selected.

We note that Mistral do not publish the size of their IFT dataset, while Meta indicates that over 10 million human-annotated samples are included in theirs. We thus restrict our analysis to Llama-3, and include Mistral in our results for completeness. 

\begin{figure}[H]
    \centering
    \makebox[\textwidth][c]{\includegraphics[width=0.8\textwidth]{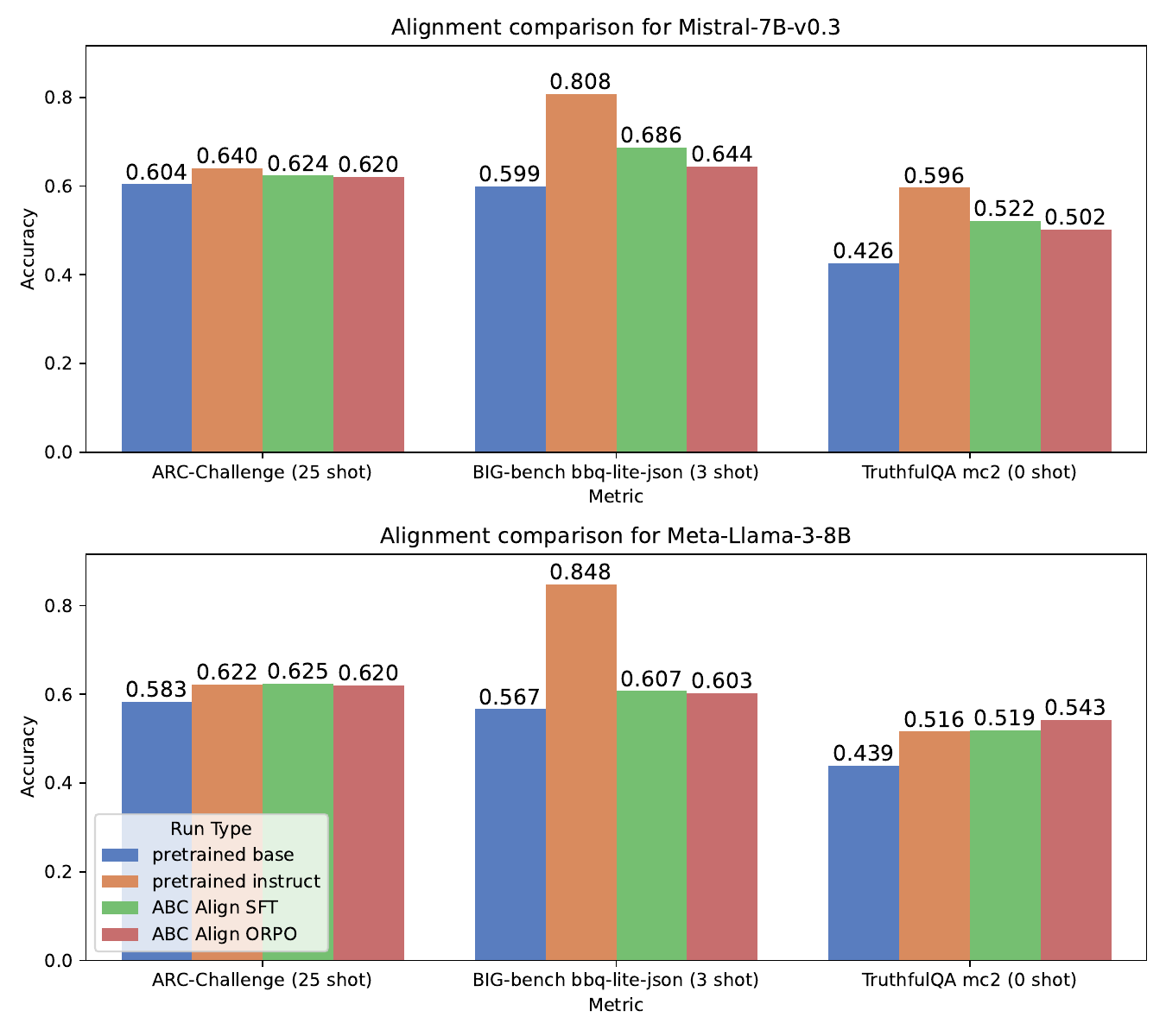}}
    \caption{Comparison between IFT and ABC Align models fine-tuned using SFT or ORPO, across \texttt{arc-challenge} (left), \texttt{bbq-lite-json} (middle) and \texttt{truthfulqa\_mc2} (right).}
    \label{fig:ift_vs_sft}
\end{figure}

We can see in Figure \ref{fig:ift_vs_sft} that our models offer near-equivalent performance gains on \texttt{arc-challenge}. Furthermore, our PO model outperforms the IFT version of Llama3 by 5.2\% on \texttt{truthfulqa\_mc2}, despite our datasets being five orders of magnitude smaller than the IFT dataset. 

The IFT model significantly outperforms both our SFT and PO models on \texttt{bbq-lite}. This result indicates that scaling dataset size and diversity of preference data is likely to offer benefits for specific alignment criteria, represented here by \texttt{bbq-lite}. Identifying socio-cultural biases is a nuanced rather than determinative task, where a wider distribution of samples would be valuable. We elaborate further on this analysis in the ‘Future Work’ section of this paper.  

\subsection{In-Context Alignment}

Here we present results of an ICA experiment on a subset of evaluations using data from our ABC Align datasets. This data includes both our internal RAG tool’s system prompt, and prompts designed specifically for the ICA setting, constructed using the ABC AI Principles. These prompts were evaluated using OpenAI’s GPT-4-turbo. It is important to note, as discussed in the ‘In-Context Alignment’ section, that these results are not comparable across the fine-tuning and ICA settings.

\begin{figure}[H]
    \centering
    \includegraphics[width=0.8\textwidth]{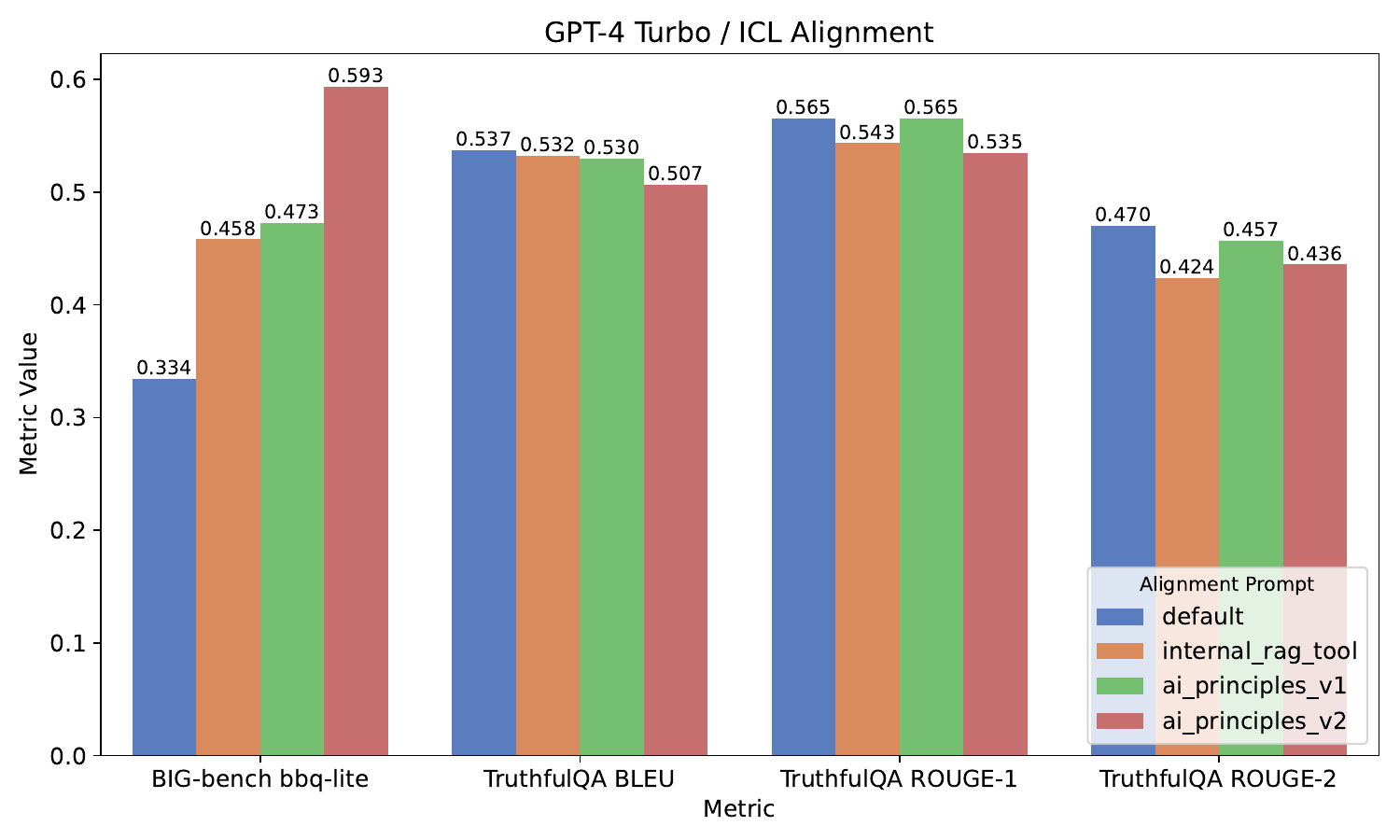}
    \caption{Evaluation of ICL Alignment using augmented prompts drawn from our internal RAG tool and the ABC AI Principles. Evaluation is conducted across \texttt{bbq-lite} and BLEU and ROUGE scores drawn from \texttt{truthfulqa\_mc2}.}
    \label{fig:icl_alignment}
\end{figure}

We do not include the prompt for our internal RAG tool in this paper. We defer it and a discussion on its construction in a future publication.
 
The prompt referred to as \texttt{ai\_principles\_v1} in Figure \ref{fig:icl_alignment} includes our AI Principles themselves, whereas \texttt{ai\_principles\_v2} is an augmentation of the former including a Q\&A pair sourced from our internal RAG tool. Please refer to Appendix B for the full v1 prompt.

\subsubsection{Evaluation on \texttt{bbq-lite}}

Compared to the baseline, the improvements across the three versions of our ICA prompts of up to 78.58\%.

\subsubsection{Evaluation on \texttt{truthfulqa\_mc2}}

Results on \texttt{truthfulqa\_mc2} demonstrate that our custom prompts reduce performance on this benchmark overall by up to 5.55\% across BLEU and ROUGE scores.

\subsubsection{Summary}

These results demonstrate that much work remains on ICA in general. We do see significant improvements across the three prompts in comparison to the baseline \textit{‘You are a helpful AI Assistant’} on the BBQ benchmark, including a 26\% improvement on BBQ for the AI Principle-based prompt that includes a canonical Q\&A pair taken from our internal RAG tool (‘v2’).  However, neither the highly-crafted internal RAG prompt nor the AI Principle-based prompt improve performance on TruthfulQA over the baseline, implying that further work exploring automated prompt optimisation and similar techniques is warranted for improved results.

\section{Discussion }
In this paper we have defined an organisational context, a PSM subject to operational constraints, that motivates our choice of models and methods. One constraint that PSMs do not face is availability of high-quality data. Indeed, that is their key competitive advantage. 

In the literature, alignment methods leverage a mixture of synthetic and human-annotated data to achieve their results, however PSMs have domain- and subject-matter expertise available at a scale that the providers of frontier models and related services do not.

Building on this insight, we have demonstrated that by applying several techniques across SFT, PO, and ICA to this data, we can outperform existing methods that use datasets constructed without that domain expertise.

Our experiments have been conducted with a small dataset, and a small number of human-reviewed samples. Integration of the underlying platform into a range of systems that support content workflows, with schema appropriate to the PO, and emerging ICA settings, will enable the scaling of our approach beyond the results presented here.

As PSMs and continue to leverage LLMs to address a variety of use-cases, our results demonstrate that they can both align these models using their existing content data, as well as editorial feedback (in the form of preferences) gathered directly from users. These data can be conditioned via natural-language prompts, which provide an additional mechanism for providing a combination of technical and editorial oversight.

Paired with evolving work on principle alignment, a key part of our contribution in this paper, organisations that implement feedback loops at multiple points of intervention in their workflows can expect to exert a measure of control over large models and their behaviours.

Evaluation of these outputs remains a key consideration. In this paper we have demonstrated the effectiveness of consistent evaluation across multiple controls, through employing the Language Model Evaluation Harness, an open, extensible framework that allows for the definition and integration of organisation-specific benchmarks.

\section{Limitations }

The aim of this paper is not to exhaustively evaluate all possible combinations of algorithm hyperparameters, model architectures and quantisation techniques. Nor has our aim been to evaluate an iterative or complete set of synthetic data and ICA prompt modifications. 

Our primary goal has been to demonstrate that a combination of methods present in the literature and our organisation’s data can be used to achieve our goal of sustainably aligning LLMs to organisational preferences and principles. We achieve this while maintaining the LLM's underlying reasoning capability while operating within organisational constrains and requirements specific to PSMs.

\section{Future Work}

In this section, we outline how we will build on the results presented here, beyond optimisation of the individual techniques used in our methodology.

\subsection{Custom Evaluations}

Defining custom evaluations, going beyond the industry-standard benchmarks we have used in this paper, is critical to ensure we are aligning LLMs towards organisational standards that are explicitly defined for measuring model capabilities and performance. We will define both general and use-case specific custom evaluations. 

\subsection{Datasets}

While we achieve results that are significant given the size of our dataset, particularly in the consistency with which our methods improve reasoning capability, comparisons with IFT versions of open-source models indicate that an increase in PO dataset size and diversity will likely yield further improvements. Following work on custom evaluations, we aim to source a variety of preference data, reflecting domain- and subject-matter expertise, to build on the results presented in this paper. The combination of methods we have outlined give us the flexibility to optimise that data for alignment of LLMs in multiple settings. 

\subsection{Prompts}

We see natural language prompts as a key point of human intervention and human oversight, both in the example of our internal RAG tool, and in defining common integration patterns and product identities, via system prompts that either guide outputs or define API integration patterns, or across a range of other applications.

The collaborative definition and evaluation of these prompts will form a key part of future work. The authors believe evolving our alignment framework to facilitate this goal is critical, particularly for emerging ‘agent’ frameworks where LLMs are a key component of goal- and task-directed systems.

In the context of a PSM, the human oversight of AI systems and the accountability \& transparency it affords will continue to be a matter of policy, compliance, and likely legislative obligation. By establishing a methodology for alignment and evaluation of LLMs, this can be incrementally improved and expanded upon as our organisation’s obligations and commitments change. We aim to enable our organisation to leverage the capabilities of new models without compromising its values or independence.

\subsection{Multimodality and ICA}

Multimodal LLMs have wide-ranging application across text, audio, and video content. We expect to evaluate and refine our methodology in this context. We consider new and emerging evidence that the embedding representation across these modalities is converging, particularly as the underlying models themselves grow larger \citep{platonic_rep_hypothesis}.

By implication, the alignment and evaluation of frontier models via ICA will be critical, and we expect to develop a methodology here that is generally applicable to any further development that accords with these early results.

\subsection{Acknowledgements}

The authors would like to thank those who have contributed to the larger body of work that this paper implies. Indeed, while this paper outlines a number of specific contributions to the Machine Learning literature, these experimental results were dependent on an infrastructure and product built by a much larger team. The team are responsible for putting in place the foundations that enable our organisation to work with LLMs at scale. Without their efforts and these foundations, our contribution and future work would not be possible.

We would like to thank the ABC’s complete ML/AI team across all domains of practice, including Scott Burns, Michael Collett, Julie Tran, Mahesh Bisl, Hanna Karlsson, Vijayan Rajendran. 

Our long-term collaboration with the ABC’s Innovation Lab has demonstrated what can be done when we work together across organisational lines. Matthew Heffernan and Andrew Hystek-Dunk worked directly on ABC Assist and were integral to the development process.

We also express our gratitude to the ABC’s Engineering, Product, and Technology leadership for the support to do this work, and our Corporate Strategy team for their invaluable feedback on ABC Assist throughout the development process.

\newpage
\section*{Appendix A}

\subsection*{ABC AI Principles}

The ABC is committed to ongoing innovation to improve the media services it delivers to audiences. AI technologies have the potential to offer new opportunities to empower content creation, strengthen research and reporting, and make ABC content more accessible. These technologies also offer ways to help our teams work more effectively. At the same time, AI technologies are not without risks and will require careful management to ensure we use these technologies responsibly and in ways that maintain or strengthen the trust that audiences place in the ABC.

Our use of AI will adhere to the following principles. We will develop policies and guidance to provide practical detail on the application of these principles.

\subsubsection*{Serving Audiences}

We will use and experiment with AI tools that can strengthen the services we provide to audiences.

\subsubsection*{Accuracy}

We will ensure that our staff are accountable for any ABC content created using AI. This is especially important for news, information and factual content.

\subsubsection*{Openness and Transparency}

We will inform our audiences about how we are using AI technologies. We will be able to explain how the AI works and how it will affect our audiences. We will not use AI in ways that could mislead our audiences.

\subsubsection*{Protecting Data}

We will work to ensure that our use of AI protects both the ABC’s data and the personal information the public entrusts to us.

\subsubsection*{Mitigating Bias}

We will seek to prevent bias in our AI data or algorithms that could perpetuate and amplify existing inequalities and lead to discrimination.

\subsubsection*{Respect for Creators' Rights}

We will always consider the rights of creators and rights-holders when using AI. We will protect our rights in the works that ABC staff create.

\subsubsection*{Experimentation and Evaluation}

We will continue to assess and experiment with AI technologies as they evolve to identify opportunities for innovation and mitigate potential risks for our audiences and organisation

\newpage
\section{Appendix B}

You are a helpful assistant working for ABC Australia and are guided by their AI Principles. 

Here is the list of principles and their definitions: 

\subsubsection*{Serving Audiences}

We will use and experiment with AI tools that can strengthen the services we provide to audiences. 

\subsubsection*{Accuracy}
We will ensure that our staff are accountable for any ABC content created using AI. This is especially important for news, information and factual content. 

\subsubsection*{Openness and Transparency}

We will inform our audiences about how we are using AI technologies. We will be able to explain how the AI works and how it will affect our audiences. We will not use AI in ways that could mislead our audiences. 

\subsubsection*{Protecting Data}

We will work to ensure that our use of AI protects both the ABC’s data and the personal information the public entrusts to us. 

\subsubsection*{Mitigating Bias}

We will seek to prevent bias in our AI data or algorithms that could perpetuate and amplify existing inequalities and lead to discrimination. 

\subsubsection*{Respect for Creators' Rights}

We will always consider the rights of creators and rights-holders when using AI. We will protect our rights in the works that ABC staff create. 

\subsubsection*{Experimentation and Evaluation}

We will continue to assess and experiment with AI technologies as they evolve to identify opportunities for innovation and mitigate potential risks for our audiences and organisation.

Here is an example of a question/answer aligned with these principles: 

... 

\newpage
\bibliographystyle{plainnat}
\bibliography{refs}

\end{document}